\documentclass[sigconf]{acmart}

\usepackage[T1]{fontenc}
\usepackage{booktabs}
\usepackage{multirow}
\usepackage{tabularx}
\usepackage{array}
\usepackage{graphicx}
\usepackage{xcolor}
\usepackage{balance}
\usepackage{enumitem}

\copyrightyear{2026}
\acmYear{2026}
\setcopyright{cc}
\setcctype{by-nc-nd}
\acmConference[SIGIR '26]{Proceedings of the 49th International ACM SIGIR Conference on Research and Development in Information Retrieval}{July 20--24, 2026}{Melbourne, VIC, Australia.}
\acmBooktitle{Proceedings of the 49th International ACM SIGIR Conference on Research and Development in Information Retrieval (SIGIR '26), July 20--24, 2026, Melbourne, VIC, Australia}
\acmISBN{979-8-4007-2599-9/2026/07}
\acmDOI{10.1145/3805712.3808463}
 
\newcolumntype{Y}{>{\raggedright\arraybackslash}X}
\newcommand{\localpct}{Local\%}

\title{Localization Boosting for Growth Markets: Mitigating Cross-Locale Behavioral Bias in Learning-to-Rank}

\author{Suryaa Veerabathiran Seran}
\orcid{0000-0002-3301-958X}
\affiliation{%
  \institution{Adobe}
  \city{San Jose}
  \state{CA}
  \country{USA}
}
\email{suryaav@adobe.com}

\author{Ashwin Naresh Kumar}
\orcid{0009-0009-5693-1350}
\affiliation{%
  \institution{Adobe}
  \city{San Jose}
  \state{CA}
  \country{USA}
}
\email{anaresh@adobe.com}

\author{Tracy Holloway King}
\orcid{0000-0002-7956-505X}
\affiliation{%
  \institution{Adobe}
  \city{San Jose}
  \state{CA}
  \country{USA}
}
\email{tking@adobe.com}

\author{Jing Zheng}
\orcid{0009-0004-6097-4146}
\affiliation{%
  \institution{Adobe}
  \city{San Jose}
  \state{CA}
  \country{USA}
}
\email{jinzheng@adobe.com}

\begin{abstract}
Adobe Express is expanding internationally, but the US  has a disproportionately large content supply and interaction volume. Learning-to-rank (LTR) models trained primarily on behavioral feedback inherit this imbalance: templates popular in US  are over-served in non-US locales. This cross-locale exposure bias suppresses local content discoverability and degrades ranking quality in growth locales.
We show that click-only training suppresses semantically informative localization features. Adding vision-language model (VLM) graded relevance labels as auxiliary supervision alongside clicks improves semantic alignment but does not preserve local content visibility. We propose a multi-objective framework combining behavioral supervision, VLM-derived relevance signals, and locale-aware boosting. Across five locales, the resulting model improves relevance while restoring stable localization, demonstrating the importance of disentangling exposure from semantic supervision.
\end{abstract}



\begin{document}
\maketitle

\section{Introduction}
Adobe Express enables users to create visual content (e.g., social media graphics) using templates. As the product expands globally, locales differ sharply in  template supply and interaction volume: the US has far more templates and substantially more behavioral data than growth markets.
Our production learning-to-rank (LTR) system is trained on click signals. While strong in mature locales, it exhibits \emph{cross-locale exposure bias}: templates popular in the US are over-scored and over-served in non-US locales, degrading ranking quality and suppressing local template discoverability.
The root cause is that clicks confound semantic relevance with historical exposure under uneven locale traffic. For example, a Japanese-language query for nightlife templates in the JP locale can surface English templates designed for the US market at top ranks because they have accumulated high global engagement, even when they are linguistically and culturally mismatched. Click-based implicit feedback is thus a biased 
proxy for relevance, shaped by presentation and position effects; non-clicks are not reliable negatives. These effects amplify historical popularity and create self-reinforcing loops that cause dominant-locale content to dominate training signals. In this setting, semantically informative localization features are systematically underweighted because they correlate with under-exposed items.

To address this, we start by introducing localization-aware multiplicative boosting in a RankNet-style objective to increase local exposure; this improves localization but can regress relevance. We then add semantic alignment features via text-embedding similarity computed from offline-generated template descriptions, but under click-only supervision these features receive low weight.
To decouple semantic relevance from behavioral bias, we use vision-language models (VLMs) to generate graded relevance labels that incorporate linguistic and cultural context. Jointly optimizing clicks with VLM supervision improves semantic feature importance and intent alignment across locales, but localization can regress because click-only learning remains implicitly popularity-weighted under exposure imbalance.
Finally, we combine click supervision, VLM-derived relevance, and calibrated locale-aware boosting in a unified multi-objective framework that balances engagement signals, semantic alignment, and local-content preference. The resulting model improves relevance while restoring stable local content visibility across locales.

This work makes the following contributions:
\begin{enumerate}[leftmargin=*,nosep]
  \item \textbf{Empirical diagnosis of exposure-induced feature suppression.} Under cross-locale imbalance, click-trained rankers overemphasize exposure-amplified signals and underweight localization features.
  \item \textbf{Semantic signal recovery via LLM supervision.} VLM-derived graded relevance labels increase the importance of text-based semantic similarity features, improving cross-locale intent alignment.
  \item \textbf{Multi-objective locale-aware ranking.} We propose a framework that combines behavioral supervision, VLM-derived relevance signals, and calibrated locale-aware boosting to improve relevance while maintaining stable localization.
  \item \textbf{Production-scale evaluation.} We evaluate across US, JP, FR, DE, and GB using offline metrics and LLM-as-a-judge evaluation.
\end{enumerate}

\section{Related Work}
\label{sec:related_work}

\textbf{Behavioral bias:}
Behavioral supervision is central to modern LTR. 
However, click logs are biased proxies for relevance due to presentation effects  \cite{joachims2005clickthrough,chapelle2009dbn,craswell2008clickmodels}. These biases  induce self-reinforcing feedback loops.
Counterfactual and unbiased LTR methods estimate propensities and optimize objectives that correct for biased feedback \cite{swaminathan2015crm,joachims2017unbiased,oosterhuis2018differentiable}, with recent work studying offline vs.\ online unbiased LTR, richer examination/observation modeling, and robustness in practical settings \cite{ai2021_ultr,chen2021_iobm,yan2022_twotower_ultr,niu2025_cltr_robustness,gupta2026_twostage_cltr}. These approaches primarily target debiasing under a single logging policy; they do not directly address cross-locale exposure imbalance where dominant-locale items receive disproportionate exposure and gradients, suppressing semantically meaningful localization signals.

\textbf{Cross-lingual and cross-locale ranking:}
Recent multilingual IR work emphasizes multilingual representation learning and dense retrieval benchmarks spanning diverse languages \cite{litschko2022_mte_clir,zhang2021_mrtydi,zhang2022_miracl}. Cross-locale ranking can be viewed as a multi-domain setting where locales differ in query intent, content supply, and behavioral volume; industrial cross-border search also exhibits severe country-level imbalance and domain-specific preferences \cite{zhang2024_damo}. These lines of work motivate modeling both semantic relevance and locale-specific objectives under distribution shift, rather than assuming a single global relevance function.

\textbf{Localization in production search systems:}
Search systems typically implement localization through a combination of (i) \emph{locale inference} (account/app locale, geolocation, and language ID), (ii) \emph{locale-aware retrieval} (per-locale indexes, region-eligibility filters, or language constraints), and (iii) \emph{ranking and blending} that incorporate locale-match features and local priors while preventing dominant-market items from overwhelming sparse locales. Empirical studies show that location materially changes result sets and can shift measured bias \cite{gezici2022_location_bias}. Geo-aware modeling has  been incorporated into query understanding (e.g., geo-aware pretraining for POI query rewriting) \cite{liu2021_geobert} and multi-country search models explicitly address country imbalance and cross-domain correlations \cite{zhang2024_damo}.

\textbf{Fairness and content balancing in ranking:}
Local content visibility is related to fairness and balancing objectives in ranked outputs. Exposure-based formulations define fairness over attention/exposure allocated by rankings \cite{singh2018fairness,biega2018equity}, with newer work highlighting challenges in exposure estimation and proposing improved stochastic/future-aware optimization \cite{heuss2022_felix,yang2023_fara}. Operationally, constraints can be enforced via post-hoc re-ranking \cite{zehlike2017fair} or at training time through objective shaping and reweighting. Our setting differs in that locality is both a user experience objective and a locale growth lever, requiring trade-offs with relevance and robustness across locales.

\textbf{LLMs for relevance labeling:}
Recent work uses large language models (LLMs) as scalable labelers for evaluation and synthetic supervision \cite{openai2023gpt4,liu2023geval,bonifacio2022inpars,dai2022_promptagator}. LLMs have also been explored as listwise rerankers \cite{sun2024_rankgpt}, while recent analyses highlight potential judge biases and interaction effects when LLMs are used to both produce and evaluate ranking outputs \cite{balog2025_rankers_judges_assistants}. In our work, LLM-derived graded relevance labels (and GPT-4o as judge) provide external semantic supervision that is less entangled with exposure \cite{openai2024_gpt4o} and increases semantic localization signals.

\section{Adobe Express Production Ranker}
We focus on five locales:
\textbf{US} (large template supply and  behavioral volume);
 \textbf{JP, FR, DE, GB} (growing, but smaller supply and  behavioral volume).
Each template is associated with a  locale of  intended audience. For evaluation, we define \textbf{local content percentage}  @\(K\).

\begin{equation}
\localpct@K = \frac{1}{K}\sum_{i=1}^{K}\mathbb{1}[\text{template}_i\ \text{is local to  request locale}]
\end{equation}

Behavioral LTR signals reflect  user preference and \emph{past exposure}. When US content dominates exposure and engagement globally, it can be treated as universally relevant by the model, even where it is culturally mismatched. This  is exacerbated when training data is pooled across locales or when non-US locales have sparse positives.
Our production baseline is a {linear SVM-rank} model trained with a {RankNet-style pairwise loss} \cite{burges2005ranknet}. For a query, let \(P=\{i: y_i=1\}\) be indices of clicked items and \(N=\{j: y_j=0\}\) indices of unclicked items. The RankNet-style pairwise loss where \(s_i = s_\theta(q,d_i)\):

\begin{equation}
\mathcal{L}_{\text{pair}}
=
\frac{1}{|P||N|}
\sum_{i\in P}\sum_{j\in N}
\log\Bigl(1+\exp\bigl(-(s_i-s_j)\bigr)\Bigr).
\label{eq:ranknet_pair}
\end{equation}


\section{Methods}
\label{sec:methods}
\label{sec:method_roadmap}

We aim to increase \emph{local discoverability}  without sacrificing  relevance: For a query issued in locale \(\ell\), we prefer locale-matching templates \emph{among similarly relevant candidates}, while avoiding  promoting irrelevant  local items.
All models use a linear scoring model \(s_\theta(q,d)=\mathbf{w}^\top \phi(q,d)\), consistent with our  SVM ranker. 

\textbf{1: Locale-aware click supervision}
\label{sec:step1_locale_click}
Starting from the click-only RankNet loss (Eq.~\ref{eq:ranknet_pair}), we make training localization-aware by \emph{reweighting} clicked-vs-unclicked pairs based on whether the clicked item matches the request locale (\S\ref{sec:locale_match}, \S\ref{sec:ranknet_locale}). This increases local exposure but regresses relevance, and the localization gains do not generalize robustly across locales and query segments.

\textbf{2: Text embeddings as a semantic localization signal.}
We add a text-embedding similarity feature computed from offline-generated template descriptions. Under click-only training, this feature receives low importance relative to exposure-amplified visual/behavioral features, limiting its impact.

\textbf{3 Multi-objective (MO): Add VLM-graded relevance supervision.}
We add graded relevance labels from a VLM and optimize a multi-objective loss combining behavioral and list-wise relevance supervision (\S\ref{sec:multiobjective}). This improves semantic relevance and increases the importance of semantic localization features, but can regress localization due to language-agnostic relevance labels.

\textbf{4 Locale-aware Multi-objective (LA-MO):  Add locale-aware boosting.}
We apply multiplicative locale boosting in  the pairwise  and list-wise  objectives so that locale-matching items receive stronger training pressure \emph{when they are  relevant} (\S\ref{sec:locale_boost}). This yields the best relevance-localization trade-off.

\label{sec:data_model}

\subsection{Supervision Signals}
\label{sec:supervision}

\textbf{Behavioral feedback (clicks).}
Behavioral supervision encodes pairwise preferences within a query list (clicked \(>\) unclicked), but is susceptible to exposure bias and can exacerbate imbalance across locales.

\textbf{Graded relevance via VLM labels.}
To inject a less biased relevance signal, we use a vision-language model (VLM) to assign an integer relevance label \(r_i \in \{0,1,2,3\}\) to each \((q,d_i)\) pair, where 3 denotes highly relevant and 0 denotes irrelevant. Labels are generated per query over the candidate set and are used as graded supervision in a list-wise objective (\S\ref{sec:listnet}). If a query list lacks VLM labels, we fall back to behavioral supervision only.

\subsection{Multi-Objective Learning-to-Rank}
\label{sec:multiobjective}

We optimize a weighted combination of a pairwise behavioral loss and a list-wise graded relevance loss. The \(\lambda\)'s control the trade-off.

\begin{equation}
\mathcal{L} = \lambda_{\text{rank}}\,\mathcal{L}_{\text{pair}} + \lambda_{\text{list}}\,\mathcal{L}_{\text{list}}.
\end{equation}

\textbf{Pairwise behavioral objective (RankNet)}
\label{sec:ranknet}
We use the RankNet-style pairwise click loss \(\mathcal{L}_{\text{pair}}\)  to encourage clicked items to be ranked above unclicked items within the query list.

\textbf{List-wise graded relevance objective (ListNet top-1)}
\label{sec:listnet}
Given VLM relevance labels \(\{r_i\}_{i=1}^{n}\) for a query list, we form a \emph{target} distribution over items using a temperature \(\tau\), model distribution from scores $q_i$ and ListNet cross-entropy $\mathcal{L}_{\text{list}}$:

\begin{equation}
p_i = \frac{\exp(r_i/\tau)}{\sum_{k=1}^{n}\exp(r_k/\tau)}.
\end{equation}

\begin{equation}
q_i = \frac{\exp(s_i)}{\sum_{k=1}^{n}\exp(s_k)}.
\end{equation}

\begin{equation}
\mathcal{L}_{\text{list}}
=
-\sum_{i=1}^{n} p_i \log(q_i).
\end{equation}

If all \(r_i\) are identical (no graded signal), we omit \(\mathcal{L}_{\text{list}}\) for that query.

\subsection{Locality Signal and Locale Match Definition}
\label{sec:locale_match}

To express locality, we associate each query with a user locale code \(\ell(q)\) and each template with a set of eligible regions \(R(d)\) (from metadata).
We define a locale-match indicator; if \(\ell(q)\) or \(R(d_i)\) is missing, we set \(m_i=0\) (no locale-specific weight).

\begin{equation}
m_i = \mathbb{1}\bigl[\ell(q)\in R(d_i)\bigr].
\end{equation}

\subsection{Locale-Aware Multi-Objective Training}
\label{sec:locale_boost}

Relevance labels can be \emph{language-agnostic} (e.g., an English template may be graded relevant for a German query), which can regress locality. To encode a soft preference for same-locale content without hard overrides, we apply \emph{multiplicative} locale boosting to both objectives.
Let \(\eta \ge 1\) be the \texttt{locale\_boost\_factor}. We apply boosting only when it reinforces an otherwise relevant local signal:
\begin{itemize}[leftmargin=*,nosep]
  \item In RankNet, we boost the weight of \emph{pairs} where a locale-matching positive should outrank a non-matching negative.
  \item In ListNet, we boost the \emph{target relevance} of locale-matching items before forming the target distribution.
\end{itemize}
In this design: (i) items labeled irrelevant (\(r_i=0\)) remain unpromoted under multiplicative boosting; (ii) locale preference is expressed as a soft bias rather than an unconditional rerank.

\subsubsection{Locale-aware RankNet via pair reweighting}
\label{sec:ranknet_locale}

We reweight each clicked-vs-unclicked pair \((i,j)\) with:
\begin{equation}
w_{ij} =
\begin{cases}
\eta, & \text{if } m_i=1 \text{ and } m_j=0,\\
1, & \text{otherwise.}
\end{cases}
\label{eq:ranknet_locale_w}
\end{equation}

The locale-aware pairwise loss becomes:

\begin{equation}
\mathcal{L}_{\text{pair}}^{\text{loc}}
=
\frac{\sum_{i\in P}\sum_{j\in N} w_{ij}\,\log\!\Bigl(1+\exp\bigl(-(s_i-s_j)\bigr)\Bigr)}
{\sum_{i\in P}\sum_{j\in N} w_{ij}}.
\label{eq:ranknet_pair_loc}
\end{equation}

This asymmetric weighting amplifies training pressure only when a locale-matching clicked item competes with a non-matching unclicked item.

\subsubsection{Locale-aware ListNet via target shaping}
\label{sec:listnet_locale}

We incorporate locale preference into the graded relevance supervision by boosting the relevance labels of locale-matching items before constructing the target distribution:
\begin{equation}
r_i' =
\begin{cases}
\eta \, r_i, & \text{if } m_i=1,\\
r_i, & \text{otherwise.}
\end{cases}
\end{equation}

\begin{equation}
p_i^{\text{loc}} = \frac{\exp(r_i'/\tau)}{\sum_{k=1}^{n}\exp(r_k'/\tau)},
\end{equation}

\begin{equation}
\mathcal{L}_{\text{list}}^{\text{loc}}
=
-\sum_{i=1}^{n} p_i^{\text{loc}} \log(q_i).
\end{equation}
Multiplicative boosting preserves the intuitive constraint that \(r_i=0\Rightarrow r_i'=0\): locale does not create relevance where none exists.

\subsubsection{Final multi-objective with locale preference}
\label{sec:final_objective}

Our final training objective is:
\begin{equation}
\mathcal{L}^{\text{final}}
=
\lambda_{\text{rank}}\,\mathcal{L}_{\text{pair}}^{\text{loc}}
+
\lambda_{\text{list}}\,\mathcal{L}_{\text{list}}^{\text{loc}}.
\end{equation}

When locale metadata is missing for a query or its items, we recover the non-locale objectives by setting \(m_i=0\) (equivalently, \(\eta=1\)).

\subsection{Curriculum Schedule for Locale Boosting}
\label{sec:curriculum}

Applying a strong locale preference too early can destabilize training, especially when non-English locales have fewer positives or noisier locale metadata. We therefore optionally use a curriculum that ramps the effective boost from 1.0 to \(\eta\) over epochs.
Let \(e \in \{1,\dots,E\}\) be the epoch index. We define:

\begin{equation}
\eta_e = 1 + \rho_e(\eta-1),
\end{equation}

\noindent where \(\rho_e \in [0,1]\) increases monotonically with training progress (e.g., linear ramp after a warm-up period). We use \(\eta_e\) in place of \(\eta\) in \S\ref{sec:ranknet_locale} and \S\ref{sec:listnet_locale}.

\section{Experiments and Results}
\label{sec:experiments}

\label{sec:variants}
\label{sec:compared_methods}

We experiment with the  methods  in \S\ref{sec:method_roadmap} and report results for the strongest variants:\footnote{We  ran {Method 1} (locale-aware clicks only) and {Method 2} (clicks + text embeddings), but both relevance and localization regressed relative to the Prod baseline.}
\begin{itemize}[leftmargin=*,nosep]
  \item \textbf{Prod baseline:} click-only RankNet-trained production ranker (no text-embedding similarity; no locale-aware boosting).
  \item \textbf{MO:} multi-objective training: clicks + VLM relevance supervision + text-embedding similarity (\(\eta=1\)).
  \item \textbf{LA-MO:} multi-objective training: clicks + VLM relevance + text-embedding similarity + multiplicative locale boosting (\(\eta>1\)).
\end{itemize}

\subsection{Datasets}
\label{sec:datasets}

\textbf{Behavioral (click) data.}
We construct training and validation instances from Adobe Express template  interaction logs. Each instance corresponds to a query impression list: a user query \(q\) issued in a locale,  with the  list of templates, their positions, and binary interaction labels. From each impression list, we form supervised preferences: clicked items are positives and unclicked items within the same shown set are negatives. This produces robust within-query supervision and avoids cross-query confounds.

\textbf{Template locality metadata.}
We join each template id to locale eligibility metadata  (\texttt{applicableRegions}). A template is  \emph{locale-matching} for a query from locale \(\ell\) if \(\ell \in \texttt{applicableRegions}(\text{ID})\). 

\textbf{Relevance (VLM-labeled) data.}
Clicks  are an imperfect proxy for relevance due to exposure bias and cold-start effects. These are amplified in lower-traffic locales where  content supply and behavioral volume are smaller. To obtain a more direct relevance signal, we sample queries and candidate templates and label each \((q,d)\) pair using a VLM. Labels are graded on a 4-level scale from irrelevant (0) to highly relevant (3). Locale context is included in the labeling prompt so that cultural and linguistic appropriateness can be reflected in the graded judgment.

\textbf{Training signal combination.}
For methods that use both supervision sources, we join behavioral query instances and VLM relevance labels using a query identifier (qid). During training, each query list contributes (i) a pairwise click loss from observed positives vs.\ negatives and (ii) a list-wise graded relevance loss when VLM labels are available for that qid. Feature vectors \(\phi(q,d)\) include production behavioral and query-dependent signals (e.g., embedding similarities, metadata match signals), and in some variants we  include a text-embedding similarity feature channel.

\subsection{Evaluation Protocol}
\label{sec:eval_protocol}

We report ranking quality and locality across \textbf{US}, \textbf{JP}, \textbf{FR}, \textbf{DE}, \textbf{GB} using held-out query lists. Because click-based metrics are confounded by exposure bias and are sparse in non-US locales, intent alignment and localized appropriateness are difficult to infer directly from behavioral logs. We therefore rely on an LLM-as-a-judge protocol to produce all reported metrics, supplemented by a behavioral locality measurement derived from template metadata.

\textbf{LLM-as-a-judge protocol.}
\label{sec:llm_judge}
We use GPT-4o as an automated judge to capture semantic relevance and locale appropriateness. The judge receives the user query, query locale, and the ranker's top-$K$ results (template title, metadata, and preview image), and scores each ranking on three rubrics, each on a continuous scale in $[0,1]$:
\begin{itemize}[leftmargin=*,nosep]
  \item \textbf{Precision@20}: proportion of retrieved templates that are semantically relevant to the query intent, allowing for thematic variations and partial matches.
  \item \textbf{Recall@20}: breadth of coverage across reasonable semantic and stylistic variations of the query topic.
  \item \textbf{Ranking Quality (NDCG@20-style)}: whether the most relevant templates appear at top positions, with an explicit bonus for locale-matching templates in the top ranks when semantic relevance is comparable.
\end{itemize}
To reduce prompt sensitivity, we run multiple prompt variants and average scores per query. 
Table~\ref{tab:rq_means} reports the averaged Ranking Quality metric; the full set of rubric scores is reported in \S\ref{sec:results_llm_metrics}.

\textbf{Locality measurement.}
Independent of the LLM judge, we measure locality directly from template metadata as \textbf{\localpct@5} and \textbf{@20}: the fraction of top-$K$ results whose eligible regions include the query's own locale. This is a within-locale measurement: for a US query, it reports the share of US-eligible content served; for a JP query, the share of JP-eligible content served. It does not measure cross-locale leakage, which is a separate phenomenon.

\textbf{Statistical significance.} Following \cite{dror-etal-2018-hitchhikers}, we test significance with the paired Wilcoxon signed-rank test \cite{wilcoxon1945} on per-query score differences, which makes no distributional assumption on bounded list-level metrics. To control the false discovery rate across regions, we apply Benjamini--Hochberg correction at $\alpha=0.05$.

\label{sec:results}

\subsection{Results: End-to-end LLM evaluation}
\label{sec:results_end2end}
\label{sec:llm_results}

We compare the {production ranker} against {LA-MO} using LLM-as-a-judge ranking quality on a stratified sample of {4.5K queries} spanning five regions (DE=899; US=806; GB=962; JP=966; FR=932). Table~\ref{tab:rq_means} reports mean ranking quality by region (higher is better).

\begin{table}[htb]
  \centering
  \caption{Mean LLM-judged ranking quality by region (Prod vs.\ LA-MO). Significance computed via paired Wilcoxon signed-rank test (one-sided, LA-MO $>$ Prod) with Benjamini--Hochberg FDR correction across regions.}
  \label{tab:rq_means}
  \setlength{\tabcolsep}{5pt}
  \begin{tabular}{lccccc}
    \toprule
    \textbf{Region} & \textbf{N} & \textbf{Prod} & \textbf{LA-MO} & \textbf{$\Delta$} & \textbf{p-value} \\
    \midrule
    US & 806 & 0.667 & \textbf{0.680}$^{***}$ & +0.012 & $<$0.001 \\
    DE & 899 & 0.661 & \textbf{0.673}$^{**}$ & +0.012 & 0.006 \\
    FR & 932 & 0.656 & \textbf{0.664}$^{*}$ & +0.009 & 0.014 \\
    GB & 962 & 0.674 & 0.682$^{\dagger}$ & +0.007 & 0.055 \\
    JP & 966 & 0.713 & 0.714 & +0.001 & 0.377 \\
    \bottomrule
  \end{tabular}
  \vspace{2pt}
  \footnotesize
  \raggedright
  Significance levels: $^{***}\,p<0.001$, $^{**}\,p<0.01$, $^{*}\,p<0.05$, $^{\dagger}\,p<0.10$ (marginal). Bold indicates statistically significant improvement at $\alpha=0.05$.
\end{table}

\noindent Overall, \textbf{LA-MO} improves mean ranking quality in all regions, with the largest gains in \textbf{US}, and \textbf{DE}, and an insignificant positive change in \textbf{JP}.

\subsection{Results: LLM-judged standard IR metrics}
\label{sec:results_llm_metrics}

To supplement the single-score ranking quality in \S\ref{sec:results_end2end}, Table~\ref{tab:region_metric_deltas} reports LLM-judged Precision@20, Recall@20, and NDCG@20 per region. We conduct this evaluation on a query set which has less than 20\% overlap in the ranked results between Prod and LA-MO. LA-MO improves all three metrics in four of five regions. The US is the only region where LA-MO does not improve NDCG, which is consistent with the observation that the production ranker was already well-tuned for US queries due to the abundance of US behavioral data; the LA-MO gains are concentrated in under-served growth locales where exposure bias has the largest effect.

\begin{table}[htb]
  \centering
  \caption{LLM-judged Precision@20, Recall@20, and NDCG@20 by region. Best value per region/metric is bolded.}
  \label{tab:region_metric_deltas}
  \setlength{\tabcolsep}{6pt}
  \begin{tabular}{llccc}
    \toprule
    \textbf{Region} & \textbf{Model} & \textbf{Prec.@20} & \textbf{Recall@20} & \textbf{NDCG@20} \\
    \midrule
    \multirow{2}{*}{DE} & Prod  & 0.78 & 0.77 & 0.64 \\
                        & LA-MO & \textbf{0.81} & \textbf{0.80} & \textbf{0.73} \\
    \midrule
    \multirow{2}{*}{US} & Prod  & \textbf{0.69} & 0.69 & \textbf{0.56} \\
                        & LA-MO & 0.67 & 0.69 & 0.55 \\
    \midrule
    \multirow{2}{*}{GB} & Prod  & 0.86 & 0.83 & 0.70 \\
                        & LA-MO & \textbf{0.88} & \textbf{0.84} & \textbf{0.72} \\
    \midrule
    \multirow{2}{*}{JP} & Prod  & 0.80 & 0.76 & 0.68 \\
                        & LA-MO & \textbf{0.84} & \textbf{0.81} & \textbf{0.71} \\
    \midrule
    \multirow{2}{*}{FR} & Prod  & 0.78 & 0.77 & 0.65 \\
                        & LA-MO & \textbf{0.82} & \textbf{0.81} & 0.65 \\
    \bottomrule
  \end{tabular}
\end{table}

\subsection{Results: Local content exposure}
\label{sec:results_locality}
\label{sec:locale_results}

We  measure whether LA-MO increases \emph{local content}. Using template metadata, we compute the \textbf{region match rate} at  \(K\in\{5,20\}\): the fraction of top-\(K\) results whose region eligibility includes the query locale code. We report  across head/torso/tail to understand how locality gains hold across query frequency.
Table~\ref{tab:region_match_rate} compares {Prod}, {MO}, and {LA-MO}. Across most locales and buckets, LA-MO achieves the highest match rate, indicating that locale boosting counteracts the tendency  to surface globally popular  templates in non-English locales. The strongest   gains appear in \textbf{DE} and \textbf{JP}. For \textbf{US}, MO reduces match rate substantially, while locale boosting restores and surpasses production. \localpct\ is a within-locale metric that measures the share of US-eligible content served for US queries, not cross-locale leakage. The cross-locale bias problem this paper addresses is that US-eligible content is over-served in non-US locales (JP, DE, FR, GB); addressing that leakage does not imply that US queries should receive fewer US-eligible results. In fact, surfacing locally eligible content is the correct behavior in every locale, including the US. In \textbf{FR}, MO barely exceeds LA-MO, suggesting that the locale boost factor may require locale-specific calibration. In \textbf{GB},  improvements are strong in  the head  but mixed in torso/tail,  indicating that boost strength interacts with query distribution.

\begin{table}[htb]
  \centering
  \caption{Region match rate (\%) for top-\(K\) results.}
  \label{tab:region_match_rate}
  \setlength{\tabcolsep}{3.5pt}
  \begin{tabular}{llcccccc}
    \toprule
    \textbf{Loc.} & \textbf{Freq.} &
    \textbf{Prod} & \textbf{MO} & \textbf{LA-MO} &    \textbf{Prod} & \textbf{MO} & \textbf{LA-MO} \\
       && \textbf{@5 } & \textbf{@5 } & \textbf{@5 } &    \textbf{@20 } & \textbf{@20 } & \textbf{@20 } \\
    \midrule
    US & head  & 46.0 & 39.8 & \textbf{54.8} & 45.9 & 38.1 & \textbf{52.6} \\
    US & torso & 44.2 & 37.9 & \textbf{50.0} & 46.9 & 38.9 & \textbf{51.2} \\
    US & tail  & 47.9 & 39.1 & \textbf{52.6} & 48.8 & 40.0 & \textbf{52.1} \\
    FR & head  & 37.4 & \textbf{40.2} & 39.8 & 35.7 & \textbf{36.5} & 35.4 \\
    FR & torso & \textbf{35.0} & \textbf{35.0} & 34.8 & \textbf{36.6} & 36.0 & 34.6 \\
    FR & tail  & 40.5 & \textbf{43.6} & 42.0 & 40.6 & \textbf{41.1} & 39.6 \\
    DE & head  & 37.4 & 41.2 & \textbf{49.2} & 36.8 & 39.5 & \textbf{46.5} \\
    DE & torso & 33.0 & 31.9 & \textbf{39.1} & 33.0 & 29.8 & \textbf{36.6} \\
    DE & tail  & 38.4 & 38.1 & \textbf{44.2} & 38.9 & 37.2 & \textbf{43.5} \\
    JP & head  & 79.4 & 81.6 & \textbf{86.6} & 78.6 & 76.7 & \textbf{81.6} \\
    JP & torso & 77.7 & 79.4 & \textbf{81.9} & 74.8 & 76.6 & \textbf{79.0} \\
    JP & tail  & 75.1 & 75.3 & \textbf{79.1} & 73.5 & 72.5 & \textbf{76.7} \\
    GB & head  & 41.9 & 33.0 & \textbf{48.8} & 38.7 & 29.4 & \textbf{39.9} \\
    GB & torso & 38.6 & 29.2 & \textbf{40.0} & \textbf{38.6} & 28.2 & 36.5 \\
    GB & tail  & \textbf{40.7} & 30.3 & 39.5 & \textbf{40.0} & 29.5 & 38.3 \\
    \bottomrule
  \end{tabular}%
\end{table}

\subsection{Qualitative  regressions}
\label{sec:results_failures}

While \textbf{LA-MO} improves overall quality and locality in most locales, the LLM analysis highlights some of {high-severity regressions}. The most prominent regressions concentrate around {animated/motion/logo} intents (e.g., FR ``logos anim\'es''), where the production ranker returns  relevant motion/logo templates but LA-MO under-ranks them. This suggests the LA-MO’s semantic supervision or locale preference does not  preserve specialized motion cues that the production model captured via historical engagement.



\section{Conclusion}
Our experiments highlight a structural pitfall in click-trained ranking systems under cross-locale imbalance: when exposure is uneven, the training objective implicitly becomes popularity-weighted, amplifying dominant-locale signals and suppressing semantically informative but under-exposed features.  
Our  multi-objective approach combines behavioral signals, VLM relevance supervision, a stronger text-based semantic localization signal, and calibrated multiplicative locale boosting to trade off relevance--localization.
Offline quality evaluations and LLM-as-a-judge analysis  capture qualitative ranking dimensions under sparse behavioral data. Overall, the final model improves relevance with stable localization in non-US locales. Our next step is to AB test the LA-MO ranker.

\section{Presenter Bio}
Suryaa Veerabathiran Seran is a Machine Learning Engineer at Adobe, where he builds end-to-end scalable multimodal search systems and large-scale model serving platforms. He works on search, discovery, and content AI, designing retrieval and multi-stage ranking systems powered by LLMs and agentic workflows. He holds a Master’s degree in Computational Data Science from Carnegie Mellon University and is passionate about advancing practical information retrieval systems.

\bibliographystyle{ACM-Reference-Format}
\balance
\bibliography{references}

\end{document}